\documentclass[10pt,twocolumn,letterpaper]{article}

\usepackage{iccv}
\usepackage{times}
\usepackage{epsfig}
\usepackage{graphicx}
\usepackage{amsmath}
\usepackage{amssymb}

\usepackage{multicol}
\usepackage{multirow}
\usepackage{arydshln}
\usepackage{overpic}
\usepackage{rotating}
\usepackage{algorithm,algpseudocode}
\usepackage{fancyhdr}

\def\ourmodel{CamDiff}
\def\etal{\textit{et al.}}

\newcommand{\figref}[1]{Fig.~\ref{#1}}
\newcommand{\secref}[1]{Sec. \ref{#1}}
\newcommand{\tabref}[1]{Tab.~\ref{#1}}

\usepackage[breaklinks=true,bookmarks=false]{hyperref}

\iccvfinalcopy 

\def\iccvPaperID{****} 

\ificcvfinal\fi

\begin{document}

\title{CamDiff: Camouflage Image Augmentation via Diffusion Model}

\author{Xue-Jing Luo $^{1\dag}$ \quad Shuo Wang$^{1,2\dag}$ \quad  Zongwei Wu $^{1}$ \quad Christos Sakaridis $^{1}$ \quad  Yun Cheng $^{1}$ \\ 
Deng-Ping Fan $^{1*}$ \quad Luc Van Gool $^{1}$
}

\maketitle
\ificcvfinal\fi

\begin{abstract}
The burgeoning field of camouflaged object detection (COD) seeks to identify objects that blend into their surroundings. Despite the impressive performance of recent models, we have identified a limitation in their robustness, where existing methods may misclassify salient objects as camouflaged ones, despite these two characteristics being contradictory. This limitation may stem from lacking multi-pattern training images, leading to less saliency robustness. To address this issue, we introduce \textbf{\ourmodel}, a novel approach inspired by AI-Generated Content (AIGC) that overcomes the scarcity of multi-pattern training images. Specifically, we leverage the latent diffusion model to synthesize salient objects in camouflaged scenes, while using the zero-shot image classification ability of the Contrastive Language-Image Pre-training (CLIP) model to prevent synthesis failures and ensure the synthesized object aligns with the input prompt. Consequently, the synthesized image retains its original camouflage label while incorporating salient objects, yielding camouflage samples with richer characteristics. 
The results of user studies show that the salient objects in the scenes synthesized by our framework attract the user's attention more; thus, such samples pose a greater challenge to the existing COD models.
Our approach enables flexible editing and efficient large-scale dataset generation at a low cost. It significantly enhances COD baselines' training and testing phases, emphasizing robustness across diverse domains. Our newly-generated datasets and source code are available at \url{https://github.com/drlxj/CamDiff}.
\end{abstract}

\section{Introduction}
\footnotetext[1]{ETH Zurich.
$^\dag$ Co-first authors.
* Corresponding author: Deng-Ping Fan (dengpfan@gmail.com).
}
\footnotetext[2]{Beijing Normal University}

\label{sec:introduction}


 \begin{figure*}
	\centering
	\begin{overpic}[width=\linewidth]{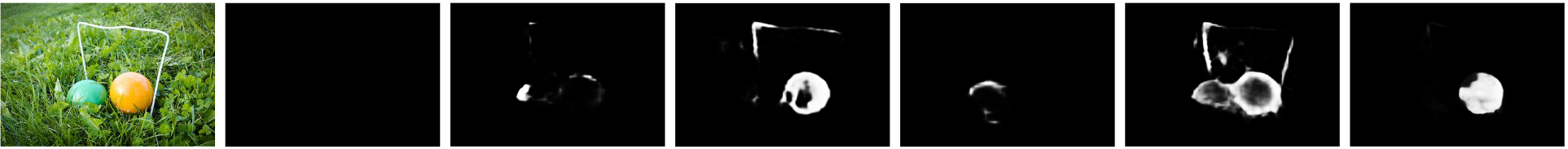}
	\small
	\put(5,-1.5){Image}
	\put(20,-1.5){GT}
	\put(32,-1.5){SINet \cite{fan2020camouflaged}}
	\put(46,-1.5){PFNet \cite{mei2021camouflaged}}
	\put(60,-1.5){C2FNet \cite{sun2021c2fnet}}
	\put(73,-1.5){SegMAR \cite{jia2022segment}}
	\put(88,-1.5){ZoomNet \cite{pang2022zoom}}
        \end{overpic}
        \vspace{2pt}
	\caption{Visualization results with current COD models tested 
 on an image with salient objects. As the object is salient, the ground truth (GT) should be all-black for the COD task. Nonetheless, the existing COD methods are less robust to the scenes with salient objects, especially the PFNet and ZoomNet.}
    \label{fig:V1}
\end{figure*}
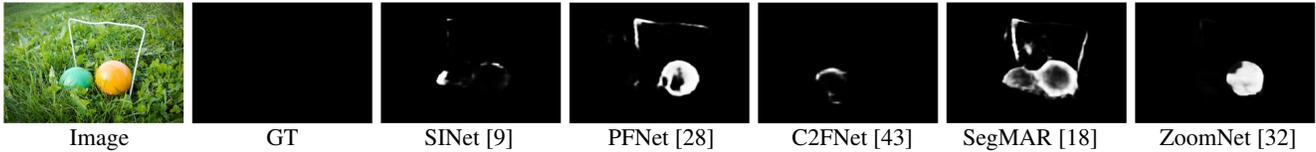

Camouflage is a predatory strategy that has evolved in natural objects through biological adaptation \cite{chu2010camouflage}. Visually, organisms alter the appearance of their bodies to match their surroundings, making them difficult to detect at first glance. Motivated by this phenomenon, a recent field of research called camouflage object detection (COD)~\cite{fan2021concealed,he2023weakly,hu2023high} has gained significant attention from the computer vision community~\cite{huang2023feature,wu2023source,zhong2022detecting}. This area of study has broad applications, including medical image diagnosis and segmentation \cite{qadir2019polyp,ucar2020covidiagnosis,dong2021polyp}, species discovery \cite{perez2012early}, and crack inspection \cite{fang2020novel}. 

In the literature, several works  \cite{fan2020camouflaged,dong2021polyp,pang2022zoom} directly extend several well-developed salient object detection (SOD) for COD tasks. However, it is noteworthy that salient and camouflaged objects are two contrasting object categories. The greater the level of saliency, the lower the degree of camouflage, and vice versa \cite{li2021uncertainty}. Hence, different strategies are imperative for detecting these two distinct object types. SOD models are based on both global and local contrasts, whereas COD models aim to avoid regions of high saliency. Unfortunately, our experiments (see \secref{sec:Experiments}) reveal a decline in the accuracy of current COD methods when both salient and camouflaged objects co-exist in an image. As Fig. \ref{fig:V1} illustrates, we tested the robustness of several state-of-the-art (SOTA) COD methods on salient objects. Nevertheless, many of these methods misclassified the salient objects as camouflaged ones. These results indicate that the current COD models are not robust enough regarding the scenes containing salient objects. 
Specifically, the algorithms employed by PFNet \cite{mei2021camouflaged} and ZoomNet \cite{pang2022zoom} detect only the more salient object (the yellow ball) and neglect the less salient object (the green ball). Thus, we speculate that existing 
COD works may only learn to distinguish the foreground and background rather than the camouflage and saliency patterns. This underscores the necessity of further research in COD to gain insight into the camouflage pattern and make COD methods truly effective.

To distinguish the saliency and camouflage patterns, one straightforward idea is to train the network via contrastive learning, which has demonstrated its effectiveness in other vision tasks \cite{dai2017contrastive, zhang2022contrastive, kang2020contragan}. As suggested in \cite{chen2020simple, tian2020makes, khosla2020supervised}, strong data augmentation can significantly benefit contrastive learning, leading to effective feature representation modeling. However, generating positive and negative pairs as contrastive samples is not feasible in our setup due to the lack of salient objects in conventional camouflage datasets. Furthermore, existing COD datasets mainly contain a single object, making the direct extension of contrastive learning infeasible. Besides, collecting and annotating a new dataset containing camouflage and salient objects within a single image would be time-consuming and labor-intensive.

In this study, we aim to enhance the robustness of future COD models regarding salient objects. To achieve this objective, we propose augmenting contrastive samples in the training data by leveraging the recent diffusion model~\cite{rombach2022high, ho2020denoising} as a form of data augmentation to generate synthetic images. This approach is inspired by the success of AI-Generated Content (AIGC)~\cite{cao2023comprehensive, ramesh2022hierarchical} and large-scale generative models. 
While some recent attempts have been to utilize diffused images for data augmentation, these efforts are only feasible for more common scenarios such as daily indoor scenes \cite{huang2023diffusion} or urban landscapes \cite{lin2023infinicity} where the domain gap is small. By contrast, we are specifically interested in camouflage scenes, which are rare and challenging for pre-trained diffusion models. These differences make our task very challenging for synthesizing multi-pattern images with large domain gaps, 
which up to our current knowledge, has not been addressed in camouflage settings.
In addition, existing works~\cite{huynh2022open} rely on additional freeze-weight deep networks to generate pseudo labels as supervision, limiting their performance and application. These limitations motivate us to design a novel framework that generates realistic salient objects within the camouflaged scenes. Our approach differs from the concurrent diffusion-augmentation methods~\cite{benigmim2023oneshot, zhang2023remodiffuse} regarding (a) the non-negligible domain gap and (b) the preserved camouflage label.

To address the target problem, in this work, we propose a diffusion-based adversarial generation framework \textbf{{\ourmodel}}. Specifically, our method consists of a generator and a discriminator. The generator is a freeze-weight Latent Diffusion Model (LDM)~\cite{rombach2022high} that has been trained on a large number of categories, making it possible to synthesize the most salient objects at scale. For the discriminator, we adopt the Contrastive Language-Image Pre-training (CLIP) \cite{radford2021learning} for its generality. Our discriminator compares the input image prompt and the synthesized object to ensure semantic consistency. To preserve the original camouflage label, we only add the generated salient object on the background, \ie, outside of the ground truth (GT) label. Therefore, our {\ourmodel} effectively transforms the problem into an inpainting task, without requiring any additional labeling cost. In such a way, we can effectively and easily enable customized editing, hence improving the development of COD from the data-driven aspect.

Our main contributions are summarized as follows:

\begin{itemize}
\item We introduce \ourmodel\, which generates salient objects on top of camouflage scenes while preserving the original label. This framework facilitates collating and combining contrastive patterns within realistic images without incurring extra costs related to learning and labeling. 

\vspace{-5pt} 
\item We conduct experiments to test the robustness of the SOTA COD methods on the COD test sets (\ie, Diff-COD), which are created from the original COD testing sets using \ourmodel. 
Our results indicate that the current COD methods are not sufficiently robust to saliency.

\vspace{-5pt} 
\item To improve the resilience of current COD methods against saliency, we generate a novel training set, called Diff-COD training set, from the original COD training sets using \ourmodel. Our experimental results demonstrate that training the existing COD models on this new training set can enhance their robustness to saliency
\end{itemize}

Overall, our research provides a fresh perspective on the notion of \emph{camouflage}, and our newly introduced camouflage synthesis tool will serve as a foundation for advancing this rapidly growing field.

\section{Related Works} 
\subsection{Diffusion Models}\label{subsec:diffsuion models}
Diffusion models \cite{ho2020denoising, rombach2022high} are generative models that generate samples from a distribution by learning to gradually remove noise from data points. Recent research \cite{dhariwal2021diffusion} shows that diffusion models outperform Generative Adversarial Networks (GANs) \cite{goodfellow2020generative} in high-resolution image generation tasks without the drawbacks of mode collapse \cite{nash2021generating} and unstable training \cite{miyato2018spectral}, and achieve unprecedented results in conditional image generation \cite{ramesh2022hierarchical}. Therefore, they have been applied in many domains such as text-to-image and
guided synthesis \cite{meng2021sdedit, poole2022dreamfusion}, 3D shape generation \cite{zhou20213d, luo2021diffusion}, molecule prediction \cite{trippe2022diffusion}, video generation \cite{yang2022diffusion} and image inpainting \cite{rombach2022high}. 

Some researchers have studied the diffusion model for image inpainting. For example, Meng~\etal~\cite{meng2021sdedit} has found that diffusion models can not only fill regions of an image but can also accomplish it conditioned on a rough sketch of the image. Another study by literature~\cite{saharia2022palette} concludes that diffusion models can smoothly fill regions of an image with realistic content without edge artifacts when trained directly on the inpainting task.

\begin{figure*}
    \centering  
    \includegraphics[width=.986\textwidth]{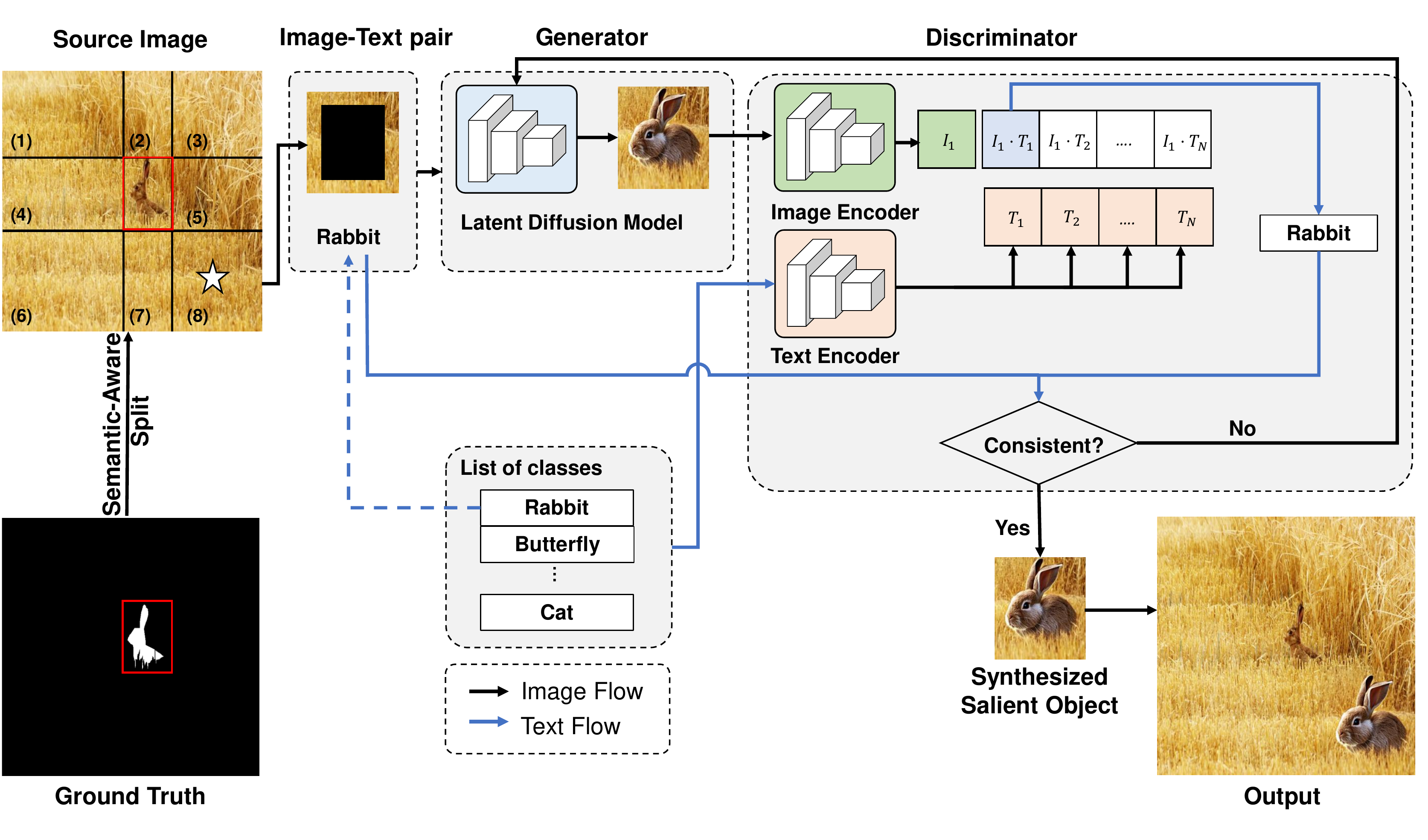}
    \vspace{-5pt}
    \caption{Our {\ourmodel} consists of a generator and a discriminator. The input of {\ourmodel} is a pair of a masked image and a text prompt. Only after the discriminator judges that the synthesized object is consistent with the text input, the synthesized image can be output and placed back into the source image. The white star in the source image means that region (8) is selected as the masked region. 
    }\label{fig:net}
\end{figure*}

\subsection{Camouflage Object Detection}\label{subsec:cod}
Camouflage object detection (COD) detects a concealed object within an image. Several research attention (\eg, SINet~\cite{fan2020camouflaged}, UGTR~\cite{yang2021uncertainty}, ZoomNet~\cite{pang2022zoom}) have focused on the comparison of COD with SOD and concluded that simply extending SOD models to solve the COD task cannot bring the desired results because the target objects have different attributes, \ie, concealed or prominent. 
To detect the concealed image, many methods have been proposed recently. For example, some methods utilize a multi-stage strategy to solve the concealment of camouflaged images. SINet \cite{fan2020camouflaged} is the first multi-stage method to locate and distinguish camouflaged objects. Another multi-stage method is SegMar \cite{jia2022segment}, which localizes objects and zooms in on possible object regions to progressively detect camouflaged objects. In addition, the multi-scale feature aggregation is the second main strategy that has been used in many methods, such as CubeNet \cite{zhuge2022cubenet}, which integrates low-level and high-level features by introducing X connection and attention fusion, as well as ZoomNet \cite{pang2022zoom}, which process the input images at three different scales to fully explores imperceptible clues between the candidate objects and background surroundings. A detailed review of COD models is out of the scope of this work; we refer readers to recent top-tier works~\cite{huang2023feature,hu2023high}.

\subsection{Camouflage Image Generation}\label{subsec:camouflage generation}
Although generating camouflage images has received limited attention, a few notable works exist in this area. One of the earliest methods, proposed in 2010, relies on hand-crafted features~\cite{chu2010camouflage}. 
Zhang \etal~\cite{zhang2020deep} have recently proposed a deep learning-based approach for generating camouflaged images. Their method employs iterative optimization and attention-aware camouflage loss to selectively mask out salient features of foreground objects, while a saliency map ensures these features remain recognizable. However, the slow iterative optimization process limits the practical application of their method.
Moreover, the style transfer of the background image to the hidden objects can often result in noticeable appearance discontinuities, leading to visually unnatural synthesized images. 
To overcome these limitations, Li~\etal~\cite{li2022location} has proposed a Location-Free Camouflage Generation Network. Although this method outperforms the previous approach~\cite{zhang2020deep} in terms of visual quality, it may fail to preserve desired foreground features or make objects identifiable using the saliency map 
in certain cases.
In summary, existing methods all follow the same strategy to produce camouflage images: They use two images to represent the foreground image and the background image respectively, and they attempt to directly integrate the foreground image with the background image by finding a place where the foreground object is hard to detect within the synthesized image.

\section{Proposed \ourmodel}\label{sec:Method}
\subsection{Overall Architecture}
To evaluate the effectiveness of existing camouflage object detection (COD) methods on negative samples (\ie, scenes with salient objects), we suggest creating synthetic salient objects on top of current camouflage datasets. Normally, when a task-specific model is trained with COD datasets, it should effectively detect the camouflaged samples, while being robust and not detecting the synthesized salient ones. Therefore, such an approach allows us to thoroughly investigate whether a learning-based COD method can accurately distinguish between camouflage and salient objects.  To achieve this objective, we propose a new generation network called \ourmodel, which is built upon existing COD datasets. Since these datasets already contain camouflaged objects with corresponding camouflage ground truth masks, 
our aim is to add synthesized salient objects into the background. 
By doing so, we can maintain the original camouflage labels and leverage them while also introducing salient samples that have contrasting characteristics. 

\figref{fig:net} illustrates the overall architecture of our proposed method. 
We start with a COD dataset, which provides us with a source image and its corresponding ground truth (GT). Using the GT, we identify the bounding box with the minimum coverage area to prevent \ourmodel~from altering the camouflaged image. Next, we divide the source image into nine areas via grid lines, using the bounding box to preserve the area where the camouflaged object is placed. 
Only eight of the areas are available for input into \ourmodel. 
We randomly select one of these regions and cut it out from the source image, covering a specific proportion (\eg, 75\% as the default setting in our experiments) of the total area from the center. We then feed the masked image into the generation network, and \ourmodel~generates a salient object within the masked area. Finally, we place the selected region back into its original location within the source image. In such a manner, we can not only preserve the GT labels for camouflaged objects but also add contradictory synthesized salient samples.

\begin{algorithm}[t!]
 	\caption{Mask generation} 
 	\begin{algorithmic}[1]
             \State Put the eight regions' index in a list $candidates$ in order
             \State Shuffle the index in $candidates$
 		\For {$i$ in $candidates$}	
                 \If{the area of region $i$ is higher than $RATIO_{MIN}$} 
                     \If{the area of region $i$ is less than $RATIO_{MAX}$} 
                         \State choose the area $mask$ that covers $RATIO_{MASK}$ of the total region area from the center
                         \State break
                     \Else
                         \State choose the area $mask$  that covers $RATIO_{MASK} \cdot RATIO_{MAX}$ of the total region area from the center
                         \State break
                \EndIf
                 \Else
                     \State continue
                 \EndIf 
			
 		\EndFor
         \State return $mask$ 
 	\end{algorithmic} 
  \label{algo}
 \end{algorithm}

\begin{table}
\small
    \centering
    \renewcommand{\arraystretch}{1.0}
     \caption{Hyperparameters setting.}
    \setlength\tabcolsep{39pt}
    \begin{tabular}{l|r}
    \hline
          Parameter       &  Value  \\
    \hline
     $RATIO_{MIN}$           & 6.25\% \\
     $RATIO_{MAX}$           & 25\%  \\
     $RATIO_{MASK}$          & 75\%  \\    
    \hline
    \end{tabular}%
  \label{tab:param}
\end{table}%

To generate the salient object, we propose a generation framework based on the Generative Adversarial Network (GAN) architecture. 
Specifically, we utilize the widely-acknowledged Latent Diffusion Model (LDM) as the generator and the Contrastive Language-Image Pre-Training (CLIP) as the discriminator. As shown in \figref{fig:net}, the input to our framework is an image with the previously-masked region, along with a text prompt that describes the target object. This masked region and text prompt are then fed into the generator. Based on the prompt, the LDM block generates the target object on top of the masked region. The filled-up region is then sent to the discriminator to determine if it matches the input prompt. If not, the generator adjusts the seed to generate a new salient object. The objective is to train the generation network to only produce validated images when the discriminator predicts a high probability of matching the input prompt.

Our framework transforms the image generation task into an inpainting task, and thus requires a mask to cover the selected region. The mask generation process is explained in Algorithm \ref{algo}. The mask is designed to cover a certain percentage of the selected region to avoid artifacts when blending the synthesized object with the source image. The ratio of the masked area to the region area is set to a constant, $RATIO_{MASK}$. The size of the selected region is crucial for the inpainting task, as it can affect the quality of the generated salient object. If the region is too small, the LDM may fill the background instead of the object, while if it is too large, the salient object may be too much larger than the concealed object, misleading COD methods. Therefore, we set an upper bound ($RATIO_{MAX}$) and a lower bound ($RATIO_{MIN}$) for the ratio between the region area and the total area of the source image. The values for these parameters are listed in \tabref{tab:param}.

\subsection{Latent Diffusion Model (LDM)}\label{subsec:LDM}
We use the LDM~\cite{rombach2022high} which is pre-trained on a large-scale dataset as our generator's base model. The LDM is a two-stage method that consists of an autoencoding model to learn the latent representation of an image and a Denoising Diffusion Probabilistic Model (DDPM)\cite{ho2020denoising}. In the first stage, the autoencoding model is trained to learn a space that is perceptually equivalent to the image space. The encoder $\mathcal{E}$ encodes the given image $x \in \mathbb{R}^{H\times W \times 3}$ to the latent representation $z \in \mathbb{R}^{H\times W \times C}$ so that $z = \mathcal{E}(x)$, while the deocder $\mathcal{D}$ reconstructs the estimated image $\Tilde{x}$ from the latent representation, such that $\Tilde{x} = \mathcal{D}(\Tilde{z})$ and $\Tilde{x} \approx x$. In the second stage, the DDPM is trained to generate the latent representation within the pre-trained latent space based on a random Gaussian noise input $z_t$. The neural backbone $\epsilon_{\theta}(z_t, t)$ of the LDM is realized as a time-conditional UNet, and the objective of the DDPM trained on latent space is simplified as:

\begin{equation}
L_{DM} := \mathbb{E}_{\mathcal{E}(x), \epsilon \sim \mathcal{N}(0, 1), t} \left[ \| \epsilon - \epsilon_{\theta}(z_t, t) \|_2^2 \right],
\label{eqn:part_1}
\end{equation}

\subsection{Conditioning LDM}\label{subsec:conditioning mechanism}
To control the image synthesis, the conditional LDM implements a conditional denoising autoencoder $\epsilon_{\theta}(z_t, y, t)$ through inputs $y$  such as text, semantic maps, or other image-to-image translation tasks \cite{rombach2022high}. The proposed \ourmodel~exploits this ability to control image synthesis through text input. To turn DDPMs into more flexible conditional image generators, their underlying UNet backbone is augmented with the cross-attention mechanism. The embedding sequences $\tau_{\theta}(y) \in \mathcal{R}^{M \times d_{\tau}}$ from the CLIP ViT-L/14 encoder is fused with latent feature maps via a cross-attention layer implementing as:
\begin{equation}
\text{Attention}(Q, K, V) = \text{softmax} \left( \frac{QK^T}{\sqrt{d}} \cdot V \right), 
\label{eqn:QKV}
\end{equation}
where $Q = W_Q^{(i)} \cdot \varphi_i(z_t), K = W_K^{(i)} \cdot \tau_\theta(y), V = W_V^{(i)} \cdot \tau_\theta(y)$, and $\varphi_i(z_t)$ is a intermediate representation of the UNet implementing $\epsilon_\theta$. 
$W_Q^{(i)}$, $W_K^{(i)}$, and $W_V^{(i)}$ are learnable projection matrix. The objective of the conditional LDM is converted from Eqn. \ref{eqn:part_1} to:

\begin{equation}
L_{CDM} := \mathbb{E}_{\mathcal{E}(x), y, \epsilon \sim \mathcal{N}(0, 1), t} \left[ \| \epsilon - \epsilon_{\theta}(z_t, t, \tau_{\theta}(y)) \|_2^2 \right],
\label{eqn:part_2}
\end{equation}

\subsection{CLIP for Zero-Shot Image Classification}\label{subsec:cascaded fusion}
To improve the quality of generated objects based on text input, it is necessary to use a discriminator that can assess the consistency of the generated objects with the text prompt. However, since the text prompt can be any arbitrary class, traditional classifiers that only recognize a fixed set of object categories are unsuitable for this task. Therefore, CLIP models offer a better option for this task.

The CLIP model comprises an image encoder and a text encoder. The image encoder can employ various computer vision architectures, including five ResNets of varying sizes and three vision transformer architectures. Meanwhile, the text encoder is a decoder-only transformer that uses masked self-attention to ensure that the transformer's representation for each token in a sequence depends solely on tokens that appear before it. This approach prevents any token from looking ahead to inform its representation better. Both encoders undergo pretraining to align similar text and images in vector space. This is achieved by taking image-text pairs and pushing their output vectors closer in vector space while separating the vectors of non-pairs. The CLIP model is trained on a massive dataset of 400 million text-image pairs already publicly available on the internet.

\section{Experiments}\label{sec:Experiments}

\begin{table}
\small
    \centering
    \renewcommand{\arraystretch}{1.3}
	\caption{Comparision of the generated dataset with the original COD and SOD dataset. The type ``orig.'' means the original dataset, while the type ``new'' means the synthesized dataset based on the corresponding COD dataset.}
    \setlength\tabcolsep{7.5pt}
    \begin{tabular}{lll|r}
    \hline
          & Dataset       &  Type   & Inception Score $\uparrow$ \\
    \hline
    \multirow{4}{*}{\begin{sideways}SOD\end{sideways}}
     & DUTSE-TE           & orig.         & 71.63\\ 
     & ECSSD              & orig.        & 24.40 \\
     & XPIE (Salient)      & orig.        & 96.79 \\
     & XPIE (Not Salient)  & orig.        & 13.96 \\
    
    \hline
    \multirow{8}{*}{\begin{sideways}COD\end{sideways}}
     & CAMO   & orig.      & 6.61 \\
     & & new         & 9.90 \\

    & CHAM & orig.   & 4.38 \\
    & & new         & 5.98 \\

    & COD10K & orig.      & 7.00 \\
    & & new         & 14.85 \\

    & NC4K   & orig.      & 7.00 \\
    & & new         & 12.87 \\
    \hline
    \end{tabular}%
  \label{tab:incep}
\end{table}%

\subsection{Experimental Setup}
\noindent\textbf{Datasets.} 
To synthesize multi-pattern images for the COD task, we selected four widely-used COD datasets: CAMO~\cite{le2019camo}, CHAM~\cite{skurowski2018chameleon}, 
COD10K~\cite{fan2020camouflaged}, 
and NC4K~\cite{lv2021simultaneously}. 
It should be noted that the COD10K dataset provides semantic labels as filenames. Therefore, we used the label directly as the text prompt. Some prompts are shown in \figref{fig:net}, which lists the classes. However, the list of classes is not directly available for the other three datasets. Since they contain common animal species such as birds, cats, dogs, \etc, we randomly chose a text prompt from the COD10K label list.

\noindent\textbf{Baselines.}
To evaluate the robustness of existing COD methods to both salient and camouflaged objects, we selected four representative and classical COD methods: 
SINet~\cite{fan2020camouflaged}, 
PFNet~\cite{mei2021camouflaged}, 
C2FNet~\cite{sun2021c2fnet}, 
and ZoomNet~\cite{pang2022zoom}, as our baselines. It is worth noting that since our paper submission, several new SOTA models have emerged, including 
FSPNet~\cite{huang2023feature} and EVP model~\cite{liu2023explicit}. 
However, this paper aims to explore new mechanisms for detecting camouflage patterns, and thus comprehensive testing of all models falls beyond the scope of this article.

\noindent\textbf{Evaluation Metrics.}
To assess the quality of the synthesized image, we employed Inception Scores~\cite{salimans2016improved}. For COD models, we follow previous works \cite{fan2021concealed,wu2023source} and evaluate the performance using conventional metrics: Mean Absolute Error ($M$), max F-measure ($F_m$), S-measure ($S_m$), and max E-measure ($E_m$).

\noindent\textbf{Implementation Details.}
Our implementation of \ourmodel~is realized in the Pytorch framework, with hyperparameters related to mask generation specified in \tabref{tab:param}. 
The whole learning process is executed on a 2080Ti GPU. 
We followed the conventional train-test split~\cite{fan2020camouflaged,fan2021concealed,zhuge2022cubenet,pang2022zoom}, using a training set of 4,040 images from COD10K and CAMO.

Among these training samples, we replaced 3,717 images with our synthesized multi-pattern images. 
The original testing samples comprised 6,473 images from CAMO, CHAM, COD10K, and NC4K. To form our Diff-COD testing set, we replaced 5,395 images with our generated images.
Although we cannot entirely replace the camouflage dataset since some images contain specific objects that the diffusion model may not generate well using the pre-trained weights, our success rate remains high. Specifically, over 92\% of the training images and 83\% of the testing images can be modified with extra salient patterns. This high success rate confirms the effectiveness of our generation framework. Note that we resized the images and masks to $512 \times 512$ to meet the requirements of the LDM.

\subsection{Quality of Synthesized Images} \label{sec:Metrics}

\begin{figure}
    \centering
\includegraphics[width=.85\linewidth]{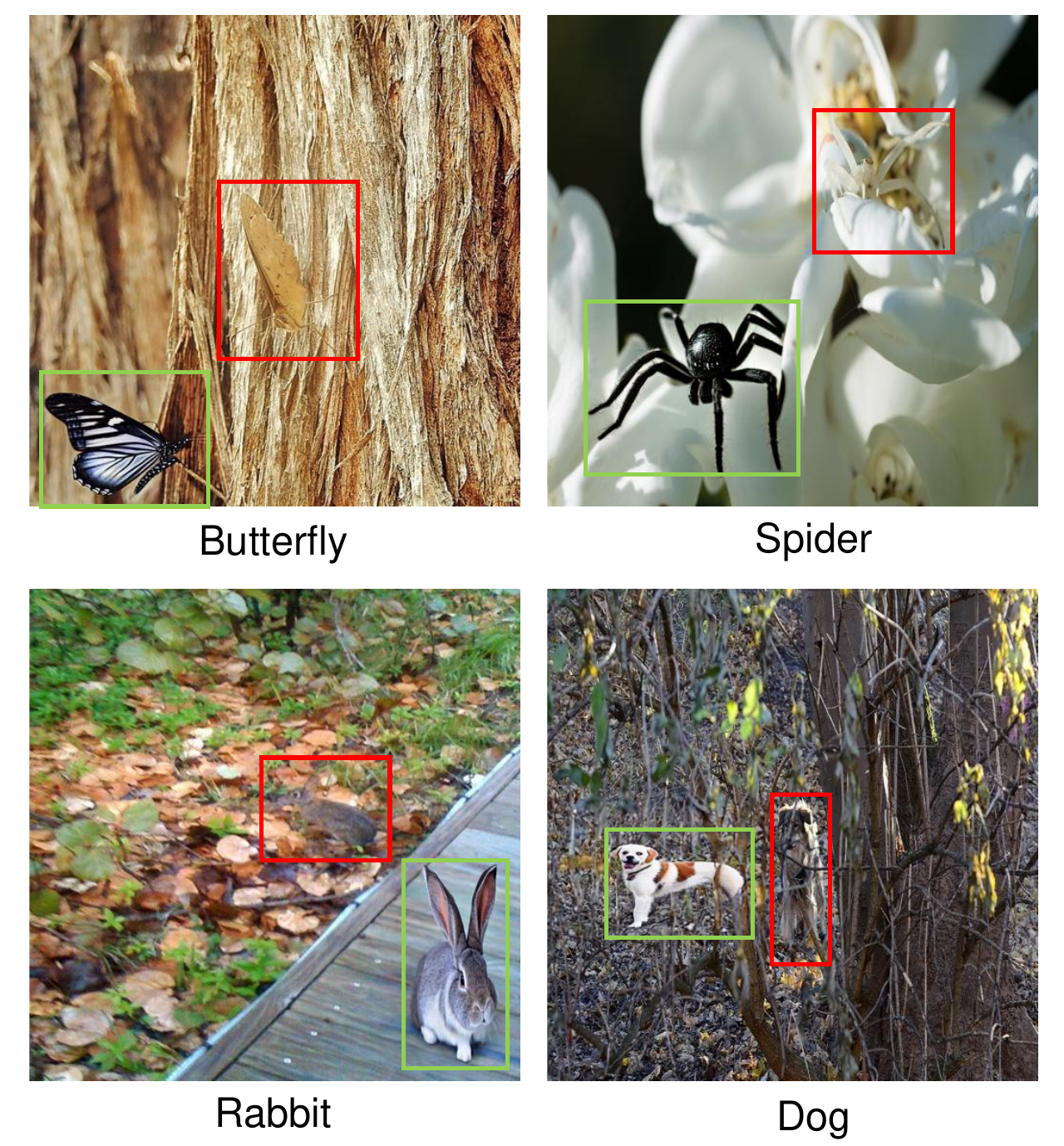}
    \vspace{-5pt}
    \caption{
    In the user study, the solution involved presenting the synthesized object within a green box, while the original object within the image was enclosed in a red box. The study results indicate that users were more likely to circle the objects in the green box, highlighting the synthesized objects as more prominent and easier to detect than the original objects within the images.
    }\label{fig:user}
\end{figure}

\noindent\textbf{Inception Score.}
To prove that our {\ourmodel} can generate a prominent object rather than a concealed object, we choose the inception score as the evaluation metric and evaluate it on the SOD datasets \cite{wang2017learning,shi2015hierarchical,xia2017and}, COD datasets \cite{le2019camo,skurowski2018chameleon,fan2020camouflaged,lv2021simultaneously}, and our generated dataset with multi-pattern images.

\tabref{tab:incep} shows that the original SOD datasets have a higher inception score than the original COD dataset, which aligns with our expectations. The rationale behind the Inception Score is that a well-synthesized image should contain easily recognizable objects for an off-the-shelf recognition system. The recognition system is more likely to detect prominent objects rather than camouflaged ones. 
As a result, images with multiple patterns tend to have a higher Inception Score than those with camouflaged patterns. By comparing the Inception Score before and after the modification, we can easily evaluate the effectiveness of our framework.
Upon replacing images in the COD dataset with multi-pattern images, it's evident that the inception score has increased across all COD datasets. This indicates that we have successfully incorporated prominent patterns on top of the original COD datasets.

\noindent\textbf{User Study.}
We conducted a user study to further evaluate the synthesized images' quality. Participants were given a subset of our synthesized images along with their corresponding labels (\eg, "Butterfly" in \figref{fig:user}) and were asked to circle the object they detected first based on the label. The salient object chosen by the user was considered the most prominent since it attracted the most human attention.
The results of our user study, with over 10 participants, showed that the average rate of users choosing the synthesized object, \ie, the salient ones, was 98\%. 
This indicates that the synthesized objects are more prominent and easier to detect than the original objects in the images.

\begin{figure}
    \centering
    \includegraphics[width=.85\linewidth]{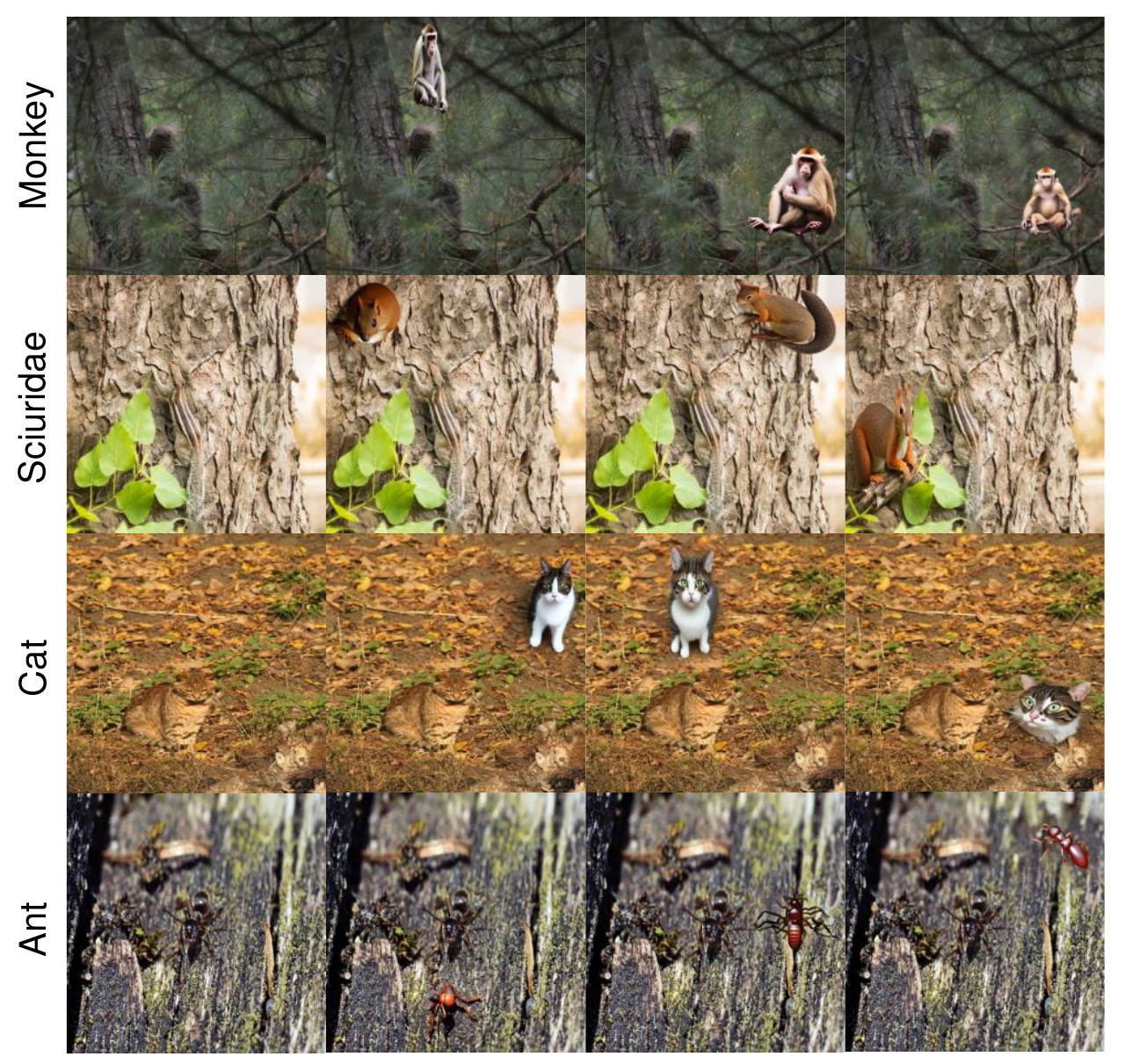}
    \vspace{-5pt}
    \caption{Examples of the synthesized images from {\ourmodel} from various classes. Each image is extended to generate three additional images of the same class, featuring objects with varying appearances. 
    }\label{fig:multi}
\end{figure}

Overall, the increased inception score and positive results from the user study support our claim that \ourmodel~generates prominent objects rather than concealed ones in the synthesized images.
In addition, \ourmodel~has demonstrated its robust capability to generate diverse objects and variations in posture for a single object type. \figref{fig:multi}~provides examples of various classes of synthesized images, each of which can be extended to generate three additional images of the same class.

\subsection{Quantitative Comparison}

In this section, we continue by introducing quantitative experiments by evaluating SOTA COD methods on the synthesized samples generated by our \ourmodel. Tab. \ref{tab:diff-cod} shows the performance of pretrained models on original and generated testing samples;   Tab. \ref{tab:cod} compares the performance trained with original COD images and our generated training samples; Tab. \ref{tab:sod} presents the robustness analysis on SOD datasets.

\begin{table}
\small
    \centering
    \renewcommand{\arraystretch}{1.3}
    \caption{Quantitative results of the pre-trained COD models on Diff-COD test dataset and COD dataset. $\uparrow$ ($\downarrow$) denotes that the higher (lower) is better. 
    }
    \setlength\tabcolsep{3.2pt}
    \begin{tabular}{ll | c | c| c| c }
    \hline

    \hline
                  \multicolumn{2}{c}{Freezed}  & SINet \cite{fan2020camouflaged} &PFNet \cite{mei2021camouflaged} & C2FNet \cite{sun2021c2fnet} &  ZoomNet \cite{pang2022zoom}\\
    \hline
\multirow{4}{*}{\rotatebox[origin=c]{90}{CAMO}} & $M \downarrow$ & .099 & .085 & .079 & .066    \\

& $F_{m}\uparrow$ & .762 & 793 & .802 & .832\\

& $S_m \uparrow$ & .751 & .782 & .796 & .819\\

& $E_m \uparrow$ & .790 & .845 & .856 & .881  \\

\hdashline

\multirow{4}{*}{\rotatebox[origin=c]{90}{Diff-CAMO}} & $M \downarrow$ & .130 & .122 & .116 & .136 \\

& $F_{m}\uparrow$ & .581 & .626 & .632 & .557 \\

& $S_m \uparrow$ & .651 & .686 & .700 & .664 \\

& $E_m \uparrow$ & .768 & .792 & .802 & .790   \\
    \hline

\multirow{4}{*}{\rotatebox[origin=c]{90}{CHAM}} & $M \downarrow$ & .044 & .033 & .032 & .023 \\

& $F_{m}\uparrow$ & .845 & .859 & .871 & .883 \\

& $S_m \uparrow$ & .868 & .882 & .888 & .900 \\

& $E_m \uparrow$ & .908 & .927 & .936 & .944  \\

\hdashline
\multirow{4}{*}{\rotatebox[origin=c]{90}{Diff-CHAM}} & $M \downarrow$ & .065 & .065 & .061 & .088 \\

& $F_{m}\uparrow$ & .700 & .795 & .726 & .596 \\

& $S_m \uparrow$ & .787 & .708 & .798 & .726 \\

& $E_m \uparrow$ & .869 & .865 & .869 & .850  \\

    \hline
    
\multirow{4}{*}{\rotatebox[origin=c]{90}{COD10K}} & $M \downarrow$ & .051 & .040 & .036 & .029  \\

& $F_{m}\uparrow$ & .708 & .747 & .764 & .799 \\

& $S_m \uparrow$ & .771 & .800 & .813 & .836 \\

& $E_m \uparrow$ & .832 & .880 & .894 & .887  \\

\hdashline

\multirow{4}{*}{\rotatebox[origin=c]{90}{Diff-COD10K}} & $M \downarrow$  & .057 & .054 & .052 & .064 \\

& $F_{m}\uparrow$ & .620 & .644 & .656 & .585 \\

& $S_m \uparrow$ & .727 & .751 & .757 & .729 \\

& $E_m \uparrow$ & .826 & .832 & .839 & .841   \\

    \hline
    
\multirow{4}{*}{\rotatebox[origin=c]{90}{NC4K}} & $M \downarrow$ & .058 & .053 & .049 & .044 \\

& $F_{m}\uparrow$ & .804 & .820 & .831 & .845 \\

& $S_m \uparrow$ & .808 & .829 & .838 & .851\\

& $E_m \uparrow$ & .873 & .891 & .898 & .896  \\

\hdashline
\multirow{4}{*}{\rotatebox[origin=c]{90}{Diff-NC4K}} & $M \downarrow$ & .090 & .084 & .080 & .076 \\

& $F_{m}\uparrow$ & .640 & .664 & .666 & .631 \\

& $S_m \uparrow$ & .719 & .744 & .746 & .739 \\

& $E_m \uparrow$ & .821 & .830 & .834 & .841   \\

    \hline
    \end{tabular}%
  \label{tab:diff-cod}
\end{table}%

\noindent\textbf{Pretrained Weights Setting.}
We created a new Diff-COD dataset to evaluate existing COD methods' effectiveness on images containing salient and camouflaged objects. This dataset includes both types of images, and we trained four SOTA COD methods (SINet~\cite{fan2020camouflaged}, PFNet~\cite{mei2021camouflaged}, C2FNet~\cite{sun2021c2fnet}, and ZoomNet~\cite{pang2022zoom}) on the Diff-COD training set. 
We then evaluated their performance on the Diff-COD testing set. It's important to note that the pre-trained LDM (low-level dense module) block can only output images with a resolution of $512\times 512$. This resolution is suitable for most existing methods trained with a resolution less than $352\times 352$.

\begin{table}
\small
    \centering
    \renewcommand{\arraystretch}{1.3}
	\caption{Quantitative results of the  test Diff-COD dataset. ``Pre.'' means the model is loaded with the pre-trained checkpoint the officially released code provides. ``Tr.'' means that the model is loaded by the checkpoints trained on our synthesized training set.}
    \setlength\tabcolsep{2.5pt}
    \begin{tabular}{ll | c c | c c | c c | c c }
    \hline 
            \multicolumn{2}{c|}{\multirow{2}{*}{Dataset}}   & \multicolumn{2}{c|}{SINet \cite{fan2020camouflaged}} & \multicolumn{2}{c|}{PFNet \cite{mei2021camouflaged}} & \multicolumn{2}{c|}{C2FNet \cite{sun2021c2fnet}} &  \multicolumn{2}{c}{ZoomNet \cite{pang2022zoom}}\\
          & & Pre. & Tr. & Pre. & Tr. & Pre. & Tr. & Pre. & Tr.\\
    \hline
\multirow{4}{*}{\rotatebox[origin=c]{90}{Diff-CAMO}} & $M \downarrow$  & .130 & .094 & .122 & .087 & .116 & .078 & .136 & .092\\

& $F_{m}\uparrow$ & .581 & .769 & .626 & .787 & .632 & .800 & .557 & .758 \\

& $S_m \uparrow$ & .651 & .753 & .686 & .773 & .700 & .789 & .664 & .773\\

& $E_m \uparrow$ & .768 & .802 & .792 & .828 & .802 & .848 & .790 & .803 \\

    \hline
\multirow{4}{*}{\rotatebox[origin=c]{90}{Diff-CHAM}} & $M \downarrow$ & .065 & .036 & .065 & .033 & .061 & .030& .088 & .058\\

& $F_{m}\uparrow$ & .700 & .864 & .795 & .858 & .726 & .870 & .596 & .764 \\

& $S_m \uparrow$ & .787 & .884 & .708 & .880 & .798 & .888 & .726 & .816 \\

& $E_m \uparrow$ & .869 & .931 & .865 & .933 & .869 & .949 & .850 & .845\\

    \hline
\multirow{4}{*}{\rotatebox[origin=c]{90}{Diff-COD10K}} & $M \downarrow$ & .057 & .047 & .054 & .041 & .052 & .038 & .064 & .053 \\

& $F_{m}\uparrow$ & .620 & .708 & .644 & .735 & .656 & .748 & .585 & .691\\

& $S_m \uparrow$ & .727 & .773 & .751 & .794 & .757 & .801 & .729 & .770 \\

& $E_m \uparrow$ & .826 & .849 & .832 & .874 & .839 & .887 & .841 & .805 \\

    \hline
\multirow{4}{*}{\rotatebox[origin=c]{90}{Diff-NC4K}} & $M \downarrow$ & .090 & .060 & .084 & .052 & .080 & .047 & .076 & .069\\

& $F_{m}\uparrow$ & .640 & .807 & .664 & .821 & .666 & .834 & .631 & .789 \\

& $S_m \uparrow$ & .719 & .811 & .744 & .830 & .746 & .840 & .739 & .814\\

& $E_m \uparrow$ & .821 & .866 & .830 & .894 & .834 & .905 & .841 & .847 \\

    \hline
    \end{tabular}%
  \label{tab:cod}
\end{table}%

\begin{table}
\small
    \centering
    \renewcommand{\arraystretch}{1.3}
     \caption{Quantitative results of the original SOD testing sets. ``Pre.'' means the model is loaded with the pre-trained checkpoint provided by the paper, 
     while ``Tr.'' means that the model is loaded by the checkpoints trained on our synthesized training set.
     }
    \setlength\tabcolsep{2.5pt}
    \begin{tabular}{ll | c c | c c | c c | c c }
    \hline
            \multicolumn{2}{c|}{\multirow{2}{*}{Dataset}}   & \multicolumn{2}{c|}{SINet \cite{fan2020camouflaged}} & \multicolumn{2}{c|}{PFNet \cite{mei2021camouflaged}} & \multicolumn{2}{c|}{C2FNet \cite{sun2021c2fnet}} &  \multicolumn{2}{c}{ZoomNet \cite{pang2022zoom}}\\
          & & Pre. & Tr. & Pre. & Tr. & Pre. & Tr. & Pre. & Tr.\\
    \hline
\multirow{4}{*}{\rotatebox[origin=c]{90}{DUTS-TE}} & $M \downarrow$  & .065 & .082 & .064 & .079 & .065 & .069 & .080 & .083  \\

& $F_{m}\uparrow$ & .820 & .760 & .808 & .748 & .807 & .780 & .715 & .718 \\

& $S_m \uparrow$ & .806 & .741 & .806 & .751 & .802 & .777 & .772 & .768 \\

& $E_m \uparrow$ & .846 & .757 & .845 & .778 & .832 & .812 & .840 & .842   \\

    \hline
\multirow{4}{*}{\rotatebox[origin=c]{90}{ECSSD}} & $M \downarrow$ & .106 & .135 & .105 & .130 & .116 & .115 & .129 & .134  \\

& $F_{m}\uparrow$ & .844 & .784 & .822 & .762 & .802 & .790 & .744 & .751 \\

& $S_m \uparrow$ & .766 & .692 & .766 & .703 & .748 & .734 & .722 & .715 \\

& $E_m \uparrow$ & .786 & .688 & .784 & .702 & .750 & .740 & .834 & .841   \\

    \hline
\multirow{4}{*}{\rotatebox[origin=c]{90}{XPIE-SAL}} & $M \downarrow$ & .090 & .119 & .093 & .115 & .099 & .101 & .115 & .123  \\

& $F_{m}\uparrow$ & .822 & .763 & .804 & .739 & .786 & .762 & .720 & .703 \\

& $S_m \uparrow$ & .770 & .691 & .762 & 697 & .749 & .728 & .723 & .705 \\

& $E_m \uparrow$ & .805 & .697 & .792 & .709 & .768 & .749 & .820 & .815   \\

    \hline
    \end{tabular}%
  \label{tab:sod}
\end{table}%

However, the current SOTA method, ZoomNet [5], requires a main resolution of $384\times384$ and an additional higher resolution with a scale of 1.5 ($576\times 576$), which is larger than the capacity of the LDM model. To ensure a fair comparison, we retrained ZoomNet with a main scale of $288\times288$ since $288\times1.5 = 432$ is less than 512 and still a relatively high resolution. To ensure equal evaluation, we trained ZoomNet on the original and our new train sets with the same main resolution of $288\times 288$.

\tabref{tab:diff-cod} compares each model's performance with its pre-trained checkpoints on both Diff-COD and original COD datasets. The results indicate that all COD methods perform significantly worse on the Diff-COD dataset. This is because these methods detect the additionally generated salient object and classify them as camouflage ones, indicating a lack of robustness to saliency. As a result, we can conclude that our Diff-COD testing set serves as a more challenging benchmark and can be used as an additional tool for robustness analysis.
\noindent\textbf{Trained on our Generated Datasets.}
As previously mentioned, our framework has the capability to generate new training samples with both salient and camouflage objects. By training on our Diff-COD dataset using only camouflage supervision, the networks should learn the distinction between the two contrasting notions and become more resilient to saliency.

\tabref{tab:cod} displays the results of the pre-trained COD models trained with original COD training sets and the newly-trained COD models on our Diff-COD training sets. It is evident that the models trained on the Diff-COD training set perform significantly better on the Diff-COD testing set compared to their counterparts.
To further confirm the effectiveness of our approach in enhancing the robustness of COD models against saliency, we conducted experiments on conventional saliency datasets, including DUTS-TE~\cite{wang2017learning}, ECSSD~\cite{shi2015hierarchical}, 
XPIE~\cite{xia2017and}. As displayed in \tabref{tab:sod}, when the models were trained using our Diff-COD dataset, their performance on saliency benchmarks declined. 
This is expected since the poorer performance on the SOD datasets indicates that the newly-trained models have truly learned the camouflage pattern but not the salient pattern. As a result, these models are better equipped to withstand the influence of salient objects.

\subsection{Qualitative Comparison}

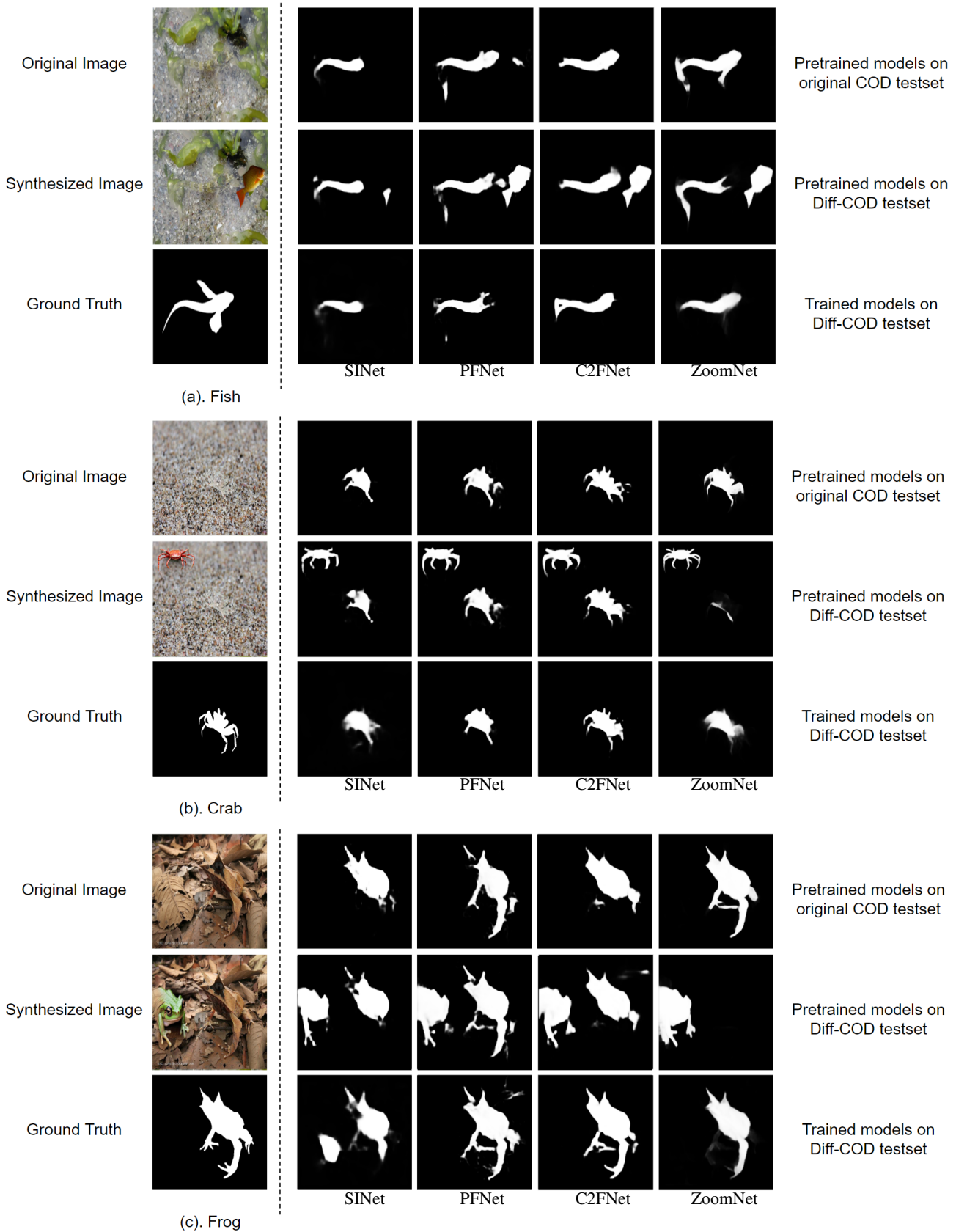
\begin{figure*}
	\centering
	\begin{overpic}[width=\linewidth]{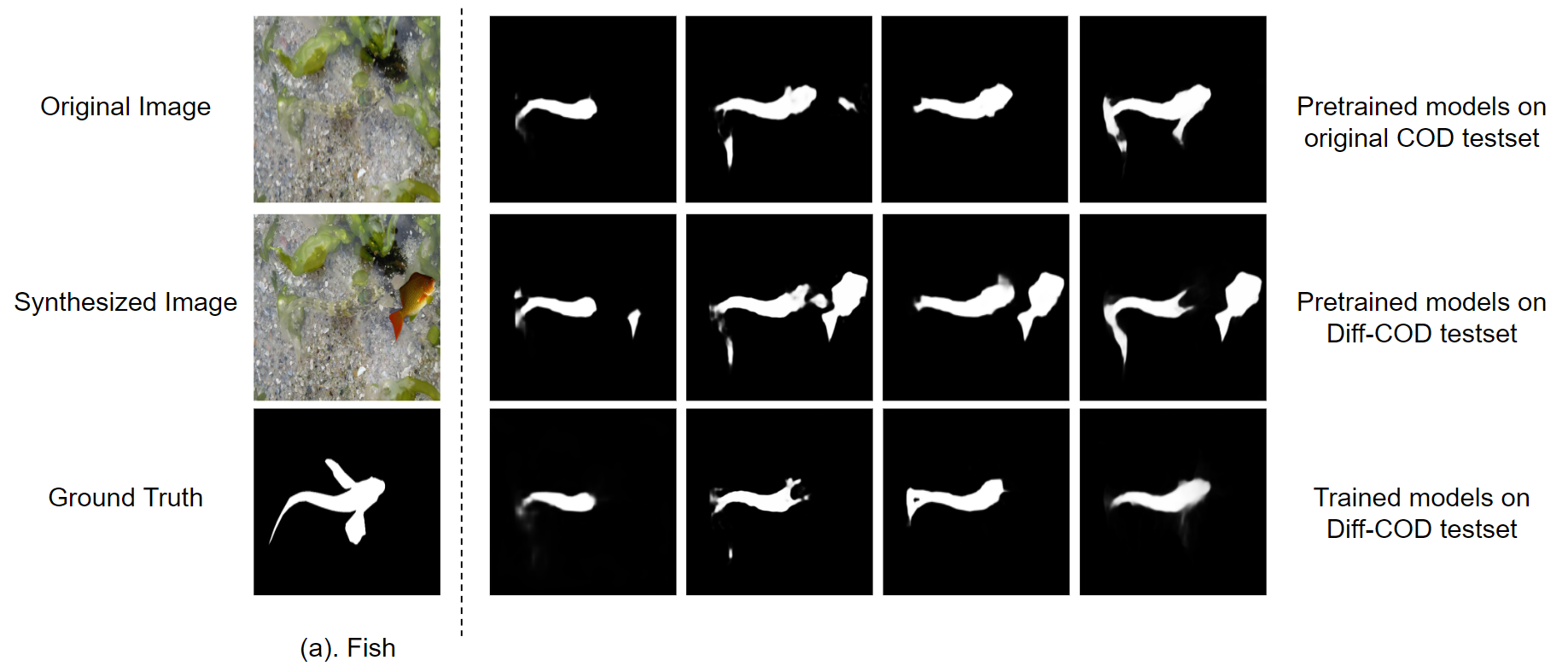}
        \end{overpic}

        \centering
	\begin{overpic}[width=\linewidth]{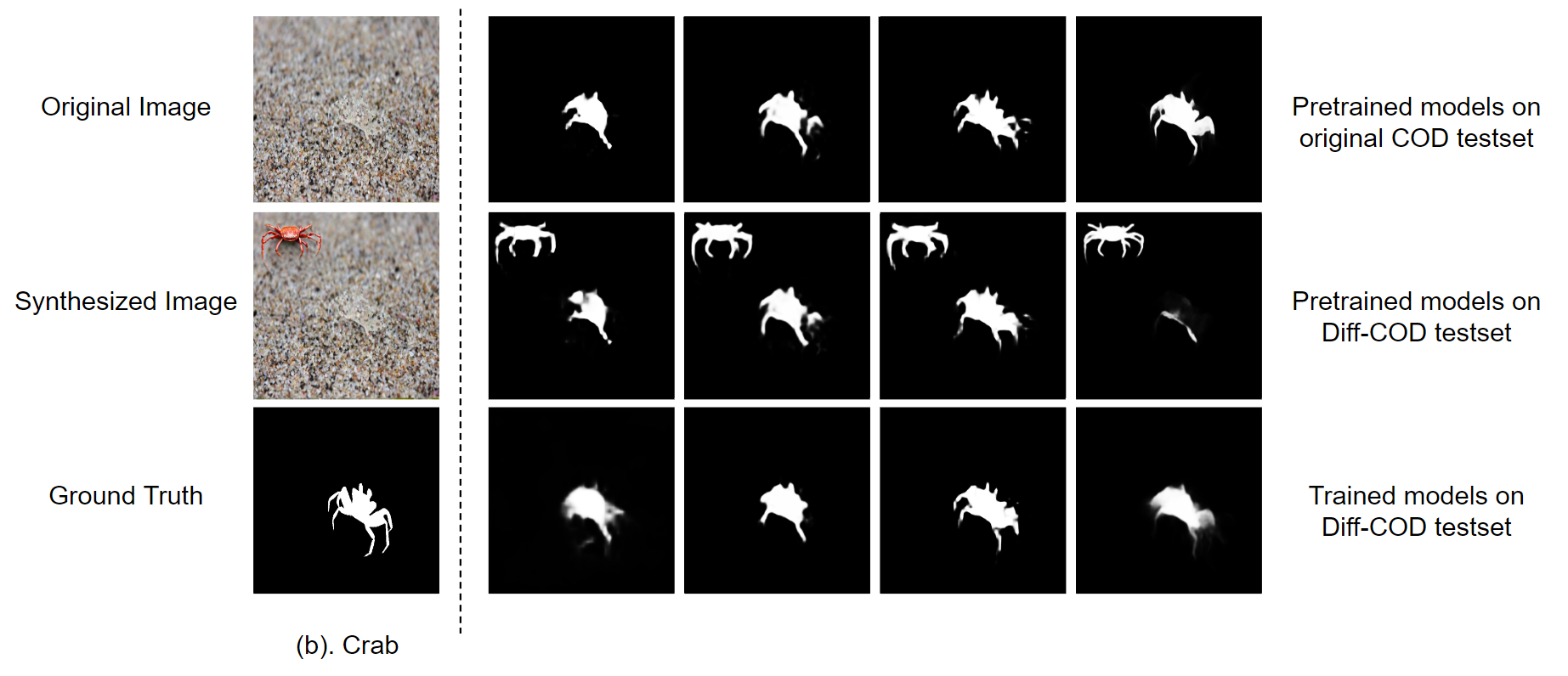}
        \end{overpic}

        \centering
	\begin{overpic}[width=\linewidth]{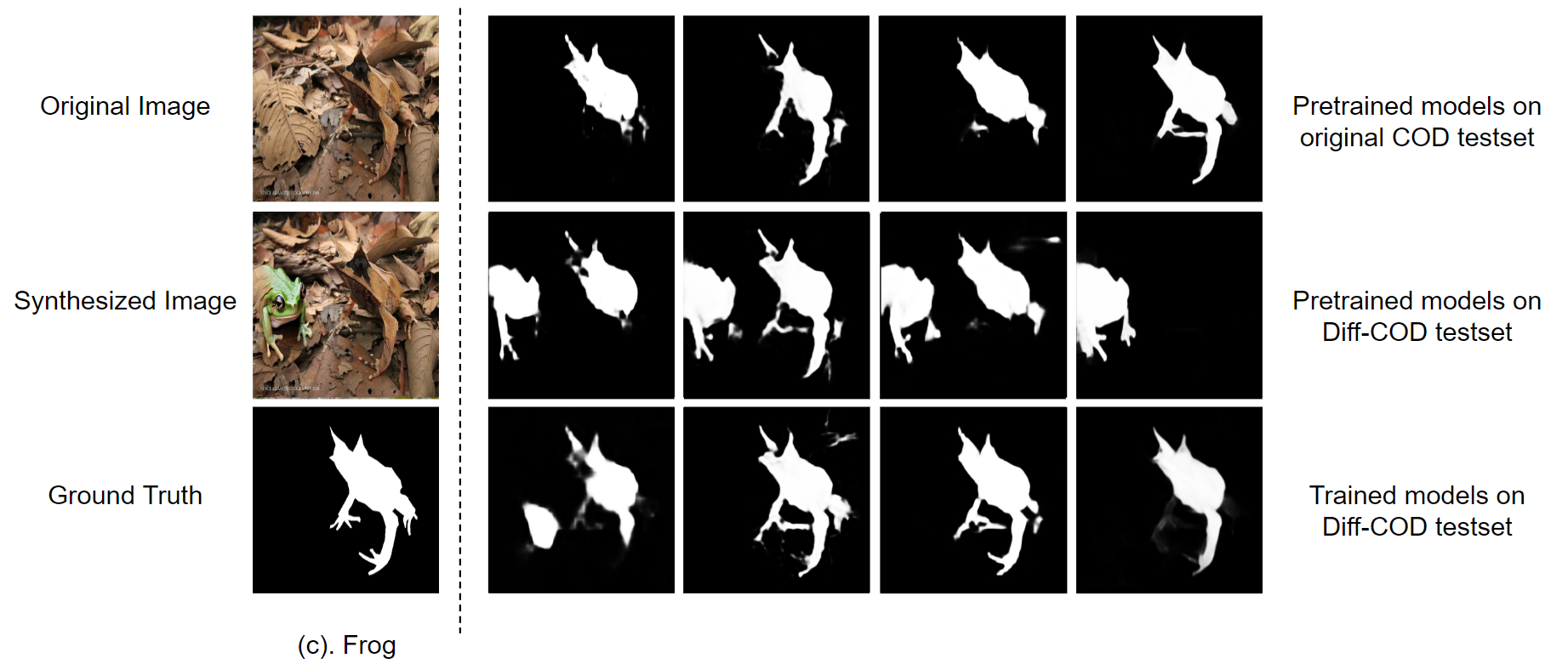}
    	\small
    	\put(36,3){SINet}
    	\put(48,3){PFNet}
    	\put(60,3){C2FNet}
    	\put(72,3){ZoomNet}
    	\put(36,46){SINet}
    	\put(48,46){PFNet}
    	\put(60,46){C2FNet}
    	\put(72,46){ZoomNet}
    	\put(36,89){SINet}
    	\put(48,89){PFNet}
    	\put(60,89){C2FNet}
    	\put(72,89){ZoomNet}

        \end{overpic}
	\caption{
        Qualitative Comparison. We conducted a qualitative comparison on three cases: Fish, Crab, and Frog. We analyzed the impact of adding salient objects to camouflaged images on pre-trained SINet, PFNet, C2FNet, and ZoomNet, respectively, by comparing the results of the first two rows. Furthermore, we evaluated the training results on the Diff-COD test set by comparing the qualitative outcomes with the pre-trained results.
        }\label{fig:compare}
\end{figure*}

\figref{fig:compare} demonstrates the effect of training on multi-pattern images on the performance of COD models. The figure is divided into three cases, each presenting the results for a different camouflaged object (fish, crab, and frog). On the left side of the dashed line in each case, the original image from the COD dataset, a synthesized multi-pattern image, and the ground truth are shown. The right side displays the results of four pre-trained models (SINet, PFNet, C2FNet, and ZoomNet) on the original COD datasets in the first row. The second row of the illustration presents the results of the models tested on the synthesized images using the same checkpoints as in the first row. Most of them detect salient objects, which is undesirable, and the accuracy of detecting camouflaged objects decreases. For instance, SINet loses some parts compared with the mask in the first row, and ZoomNet ignores camouflaged objects. These results indicate that COD methods lack robustness to saliency.
The third row of the illustration presents the results of the models trained on our Diff-COD dataset and then tested on the synthesized images. Compared to the second row, the robustness to saliency improves significantly. Nevertheless, compared to the first row, ZoomNet loses some parts of the camouflaged object. We believe this may be caused by adding noise in the training set making the fitting more difficult, but we plan to evaluate the cause in future work.

Overall, it can be concluded from \figref{fig:compare} that the presence of salient objects harms the performance of COD models in detecting camouflaged objects. However, training the COD models on multi-pattern images increases their robustness to the effects of salient objects.

\section{Conclusion}


In summary, our work introduces \ourmodel, a framework that generates salient objects while preserving the original label on camouflage scenes, enabling the easier collation and combination of contrastive patterns in realistic images without incurring extra costs related to learning and labeling. Through experiments conducted on Diff-COD test sets, we demonstrate that current COD methods lack robustness to negative examples (\eg, scenes with salient objects). 
To address this limitation, we create a novel Diff-COD training set using \ourmodel. Our experimental results demonstrate that training existing COD models on this set improves their resilience to saliency. Overall, our work provides a new perspective on camouflage and contributes to the development of this emerging field.


\noindent\textbf{Future Work.} 
We aim to extend our framework to consider original images with multiple objects and save room for their generation. Additionally, while we only implemented multipattern images as the data augmentation method in our experiments, we plan to evaluate the results using other data augmentation methods to provide a more comprehensive analysis of the impact of multi-pattern images on the performance and robustness of these models.


{\small
\bibliographystyle{ieee_fullname}
\bibliography{iccv2023AuthorKit/Reference}
}

\end{document}